\documentclass{article} 
\usepackage{iclr2027_conference,times}

\usepackage{amsmath,amsfonts,bm}

\def\eqref#1{equation~\ref{#1}}

\def\1{\bm{1}}

\DeclareMathAlphabet{\mathsfit}{\encodingdefault}{\sfdefault}{m}{sl}
\SetMathAlphabet{\mathsfit}{bold}{\encodingdefault}{\sfdefault}{bx}{n}

\usepackage{hyperref}
\usepackage{url}
\usepackage{booktabs,multirow,bm,graphicx}

\title{Reduce, Then Encode: Multiscale Volumetric Reduction for 2D Foundation Models in Brain MRI}

\author{Dexuan Ding \\
Macquarie University\\
\And
Yuankai Qi \thanks{Corresponding author}  \\
Macquarie University \\
\texttt{\{yuankai.qi\}@mq.edu.au} \\
\And
Bogong Wang \\
Australian National University \\
\AND
Luping Zhou \\
University of Sydney \\
\And
Amin Beheshti \\
Macquarie University \\
}

\newcommand{\trainableicon}{%
    \includegraphics[height=2.5ex]{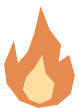}%
}

\newcommand{\frozenicon}{%
    \includegraphics[height=2.5ex]{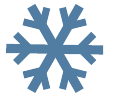}%
}

\iclrfinalcopy 
\begin{document}

\maketitle

\begin{abstract}
Pretrained 2D foundation models offer a practical alternative to dedicated 3D pretraining for brain structural magnetic resonance imaging (sMRI), but their use on volumetric data requires bridging the mismatch between a 2D encoder and a 3D volume input. Existing methods typically encode slices independently and integrate their features afterwards. We introduce Multiscale Volumetric Reduction (MVR), a reduce-then-encode approach that compresses each anatomical view from \(D\) slices into \(M \ll D\) complementary 2D components before foundation-model encoding. MVR combines an uncentered-PCA base component derived from the original through-plane intensities with residual detail components constructed from multiscale spatial descriptors. The reduction is estimated from the training volumes without diagnostic labels or gradient-based optimization and remains fixed thereafter. The resulting components are independently processed by a shared frozen 2D foundation model and concatenated for linear probing. Under this frozen-encoder setting, MVR achieves strong overall performance across ADNI, OASIS, and ABIDE relative to the evaluated 2D-to-3D adaptation methods and simple input-reduction baselines, while also generalizing strongly from ADNI to AIBL.
\end{abstract}

\section{Introduction}

Brain structural magnetic resonance imaging (sMRI) provides a non-invasive view of brain anatomy and is widely used to study structural alterations associated with neurological and neurodevelopmental disorders~\citep{BrainIAC,ENIGMA_ASD}. Unlike ordinary 2D images, an sMRI scan is represented as a three-dimensional volume, where anatomical and disease-related variations may extend across multiple slices and spatial scales. Effective analysis therefore requires preserving informative volumetric structure while extracting features useful for downstream prediction.

\begin{figure}[tbp]
    \centering
    \includegraphics[width=0.95\linewidth]{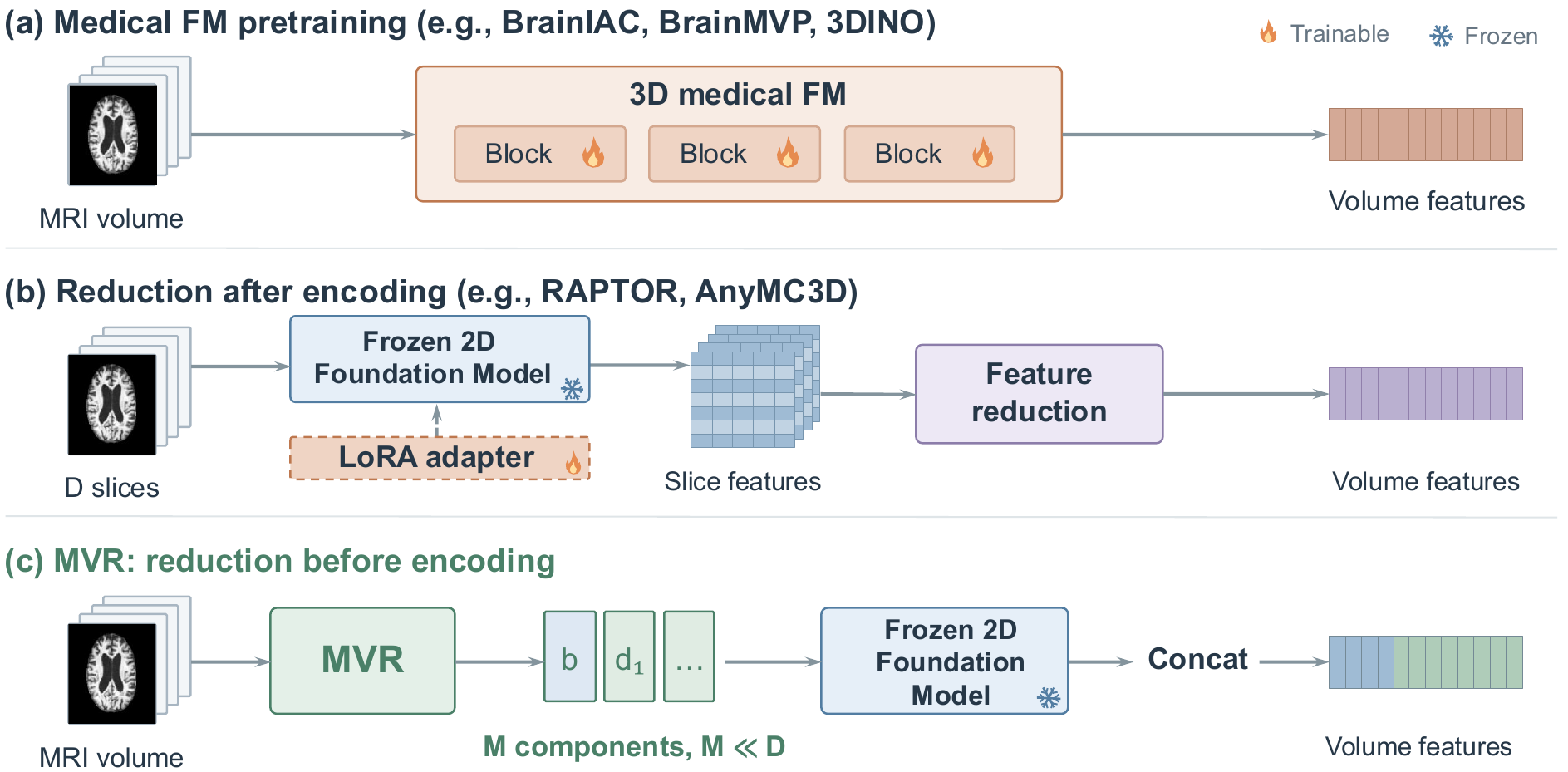}
    \caption{
    Comparison of volumetric representation strategies for brain MRI.
    (a) Medical foundation models learn volumetric representations through dedicated 3D pretraining.
    (b) Approaches based on pretrained 2D foundation models encode 
    $D$ selected 2D
    slices 
    independently and integrate the resulting features after encoding, with method-specific adaptation where applicable.
    (c) Our MVR instead performs reduction in the input space, transforming each volume into $M\ll D$ dense 2D components before encoding with a shared frozen 2D foundation model. The resulting component features are concatenated to form the volume representation. \trainableicon~and \frozenicon~denote trainable and frozen modules.}
    \label{fig:fig1}
\end{figure}

A direct solution is to learn representations from 3D medical volumes themselves. Recent medical foundation models such as BrainIAC~\citep{BrainIAC}, BrainMVP~\citep{BrainMVP}, and 3DINO~\citep{3DINO} demonstrate the effectiveness of large-scale volumetric pretraining, but require dedicated 3D architectures and substantial medical imaging data (Fig.~\ref{fig:fig1}(a)). An attractive alternative is to reuse general-purpose 2D vision foundation models pretrained on large and diverse image collections. Although their pretraining domain differs substantially from medical imaging, recent studies have shown that their representations can still transfer effectively to volumetric medical imaging~\citep{DinoAtten3D}. RAPTOR~\citep{RAPTOR}, for example, constructs volumetric embeddings using a frozen 2D foundation model (DINOv2~\citep{dinov2}), while AnyMC3D~\citep{AnyMC3D} further improves transfer through lightweight adaptation and learned aggregation. These results suggest that general-purpose 2D foundation models can provide useful representations for medical volumes without requiring dedicated volumetric pretraining.

However, a remaining challenge is the mismatch between a 2D encoder and a 3D volume. A 2D foundation model processes one two-dimensional input at a time (Fig.~\ref{fig:fig1}(b)), whereas an MRI scan contains a spatially ordered sequence of slices. Existing approaches commonly address this mismatch through an encode-then-integrate strategy: slices are first encoded independently, after which their features are pooled, projected, attended to, or otherwise aggregated into a volume-level representation. RAPTOR~\citep{RAPTOR} reduces independently extracted slice features through pooling and random projection, while AnyMC3D~\citep{AnyMC3D} combines slice-wise encoding with learned feature aggregation. 
In this paradigm, cross-slice information is incorporated only after slice-level feature extraction. When the 2D encoder is frozen, however, its slice-level representations cannot adapt to facilitate subsequent volumetric aggregation. This raises the possibility that preserving cross-slice structure in the encoder inputs themselves may provide a more effective inductive bias than relying exclusively on post-encoding integration. This motivates us to instead integrate volumetric information before encoding.

To this end, we introduce Multiscale Volumetric Reduction (MVR), a reduce-then-encode approach that transforms each anatomical view from \(D\) slices into \(M \ll D\) dense 2D components before foundation-model encoding (Fig.~\ref{fig:fig1}(c)). The key challenge is to achieve substantial through-plane reduction without collapsing useful volumetric variation. Simple averaging can suppress localized structure, while a single linear projection provides only one summary of the through-plane intensity pattern. MVR therefore adopts a base-and-detail construction. A base component projects the original through-plane intensities onto a shared direction estimated by uncentered PCA, providing a compact reference derived from the original volume. To retain complementary information beyond this single projection, MVR constructs multiscale descriptors from differences between Gaussian-smoothed slices, incorporating in-plane context at multiple spatial scales while preserving the complete through-plane sequence. After removing variation linearly associated with the base component, PCA of the residual descriptors produces \(M-1\) detail components capturing dominant complementary variation. 
The reduction operators are data-adaptively estimated from the training volumes without diagnostic labels or gradient-based optimization and remain fixed thereafter.
Each component is independently processed by the same frozen 2D foundation model, and the resulting features are concatenated across components and anatomical views to form the final volumetric representation.

We evaluate MVR on Alzheimer’s disease classification using ADNI and OASIS-3, autism classification using ABIDE, and external ADNI-to-AIBL evaluation. With a frozen DINOv3 encoder and a linear classifier, MVR achieves strong overall performance relative to the evaluated 2D-to-3D adaptation methods and simple input-reduction baselines across ADNI, OASIS, and ABIDE. Ablation studies further show that the multiscale construction and the combination of base and detail components contribute consistently to classification performance.

Our main contributions are threefold:
\begin{itemize}
\vspace{-0.5em}
    \item We introduce a reduce-then-encode strategy for adapting frozen 2D foundation models to volumetric brain MRI, incorporating information across slices before rather than only after 2D feature extraction.
    \item We develop MVR, a data-adaptive and gradient-free base-and-detail reduction that converts \(D\) slices into \(M \ll D\) complementary 2D components by combining through-plane reduction with multiscale in-plane context.
    \item We demonstrate the effectiveness of MVR across multiple brain-MRI cohorts and diagnostic tasks, including external ADNI-to-AIBL evaluation, and show through ablation that its multiscale construction and base-detail decomposition consistently contribute to classification performance.
\end{itemize}

\section{Related Work}

\paragraph{Medical foundation models.}
Medical pretraining has been used to learn transferable representations for volumetric imaging. BrainIAC learns generalizable representations from unlabeled brain MRI~\citep{BrainIAC}, while BrainMVP uses multi-parametric MRI with reconstruction, contrastive learning, and distillation~\citep{BrainMVP}. 3DINO extends self-supervised pretraining to large-scale multi-organ 3D medical data~\citep{3DINO}. These approaches obtain volumetric representations through pretraining directly on 3D medical images. In contrast, we consider the complementary setting in which a general-purpose pretrained 2D foundation model is kept frozen, and ask how volumetric information should be represented for effective 2D encoding.

\begin{figure}[tbp]
    \centering
    \includegraphics[width=1\linewidth]{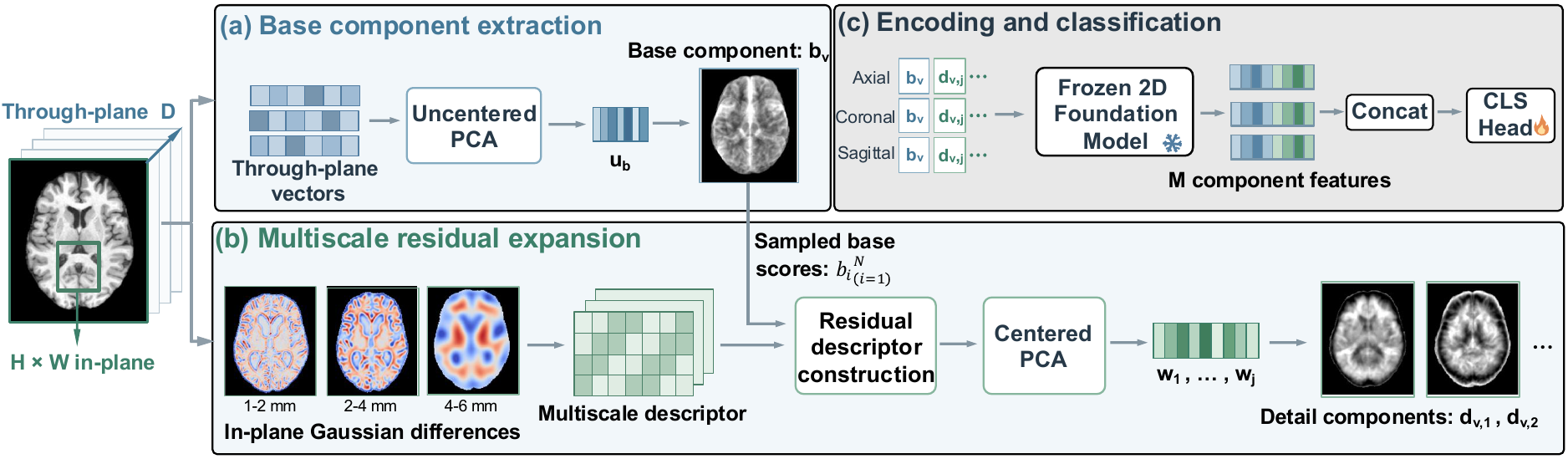}
    \caption{Main architecture of our Multiscale Volumetric Reduction (MVR). (a) Base component extraction summarizes the original through-plane intensities into a dense base component and sampled base scores. (b) Multiscale residual expansion incorporates in-plane variation at multiple scales and produces $M-1$ complementary detail components after accounting for the base reference. (c) Encoding and classification applies MVR to the axial, coronal, and sagittal views, independently encodes the resulting components with a shared frozen 2D foundation model, and concatenates their features for classification.}
    \label{fig:mvr_pipeline}
    \vspace{-1em}
\end{figure}

\paragraph{Adapting 2D models to medical volumes.}
Pretrained 2D foundation models can be extended to volumetric data by aggregating features extracted from individual cross-sections. RAPTOR reduces frozen 2D features through pooling and random projection~\citep{RAPTOR}, while AnyMC3D combines lightweight encoder adaptation with learned aggregation across slices and views~\citep{AnyMC3D}. Both perform volumetric integration after 2D feature extraction. MVR instead integrates cross-slice information in the input space before encoding. Other related methods, such as Eigenslices~\citep{Eigenslices}, reduce 3D brain MRI to a small set of 2D projections, but rely on a trainable 2D CNN to adapt to the projected representation.

\section{Method}

\subsection{Overview}
\label{sec:task}
\vspace{-0.6em}

Let $I$ denote a normalized brain sMRI volume. We denote the axial, coronal, and sagittal views by $\mathcal{V}=\{a,c,s\}$. For each anatomical view $v\in\mathcal{V}$, we use $I_v\in\mathbb{R}^{H\times W\times D}$ for the reoriented volume, where $H$ and $W$ denote the in-plane dimensions and $D$ denotes the number of slices.

Given $I_v$, MVR constructs $M\ll D$ dense 2D components before feature extraction by the pretrained 2D foundation model. A base component summarizes the original through-plane intensities using uncentered PCA (Sec.~\ref{sec:base}), while $M-1$ detail components capture complementary multiscale spatial variation after removing variation associated with the base reference (Sec.~\ref{sec:detail}). The resulting components are encoded independently by a shared frozen 2D foundation model, and their features are concatenated across components and anatomical views to form the final volume representation for downstream classification (Sec.~\ref{sec:training}).

\vspace{-1em}
\subsection{Base Component Extraction}
\label{sec:base}

As illustrated in Fig.~\ref{fig:mvr_pipeline}(a), MVR first constructs a base component that summarizes the original intensities across the slice sequence. At each in-plane location $(h,w)$, the through-plane vector $\mathbf{x}_v(h,w)\in\mathbb{R}^{D}$ contains the intensities $I_v(h,w,d)$ in slice order, for $d=1,\ldots,D$. A shared projection (detailed below) combines these $D$ entries into one base-component value while retaining its corresponding in-plane location.

We fit this projection using through-plane vectors sampled from the training volumes. From each of the $S$ training subjects, we uniformly sample $P$ vectors containing at least one intensity value different from the volume minimum. Let $N=SP$ denote the total number of sampled vectors and $\mathbf{x}_i\in\mathbb{R}^{D}$ denote the $i$-th sampled vector. Stacking them as rows gives the training matrix
$X_v=[\mathbf{x}_1,\ldots,\mathbf{x}_N]^\top\in\mathbb{R}^{N\times D}$.

The projection vector $\mathbf{u}_b\in\mathbb{R}^{D}$ is then obtained by applying uncentered PCA to $X_v$. Specifically, $\mathbf{u}_b$ is the unit eigenvector corresponding to the largest eigenvalue of the second-moment matrix $M_b=X_v^\top X_v/N\in\mathbb{R}^{D\times D}$. Unlike centered PCA, uncentered PCA retains the average intensity structure across slice positions as part of the projection. The selected direction therefore captures the dominant through-plane intensity pattern in the sampled vectors.

Once obtained, $\mathbf{u}_b$ remains fixed and serves two purposes. Applying it to the sampled vectors gives the base scores used for residual fitting,
$(b_1,\ldots,b_N)^\top=X_v\mathbf{u}_b\in\mathbb{R}^{N\times 1}$,
where $b_i=\mathbf{x}_i^\top\mathbf{u}_b$. Applying the same projection at every in-plane location of a volume gives
\begin{equation}
b_v(h,w)
=
\mathbf{u}_b^\top\mathbf{x}_v(h,w)
=
\sum_{d=1}^{D}u_{b,d}I_v(h,w,d),
\end{equation}
where $u_{b,d}$ is the weight assigned to slice $d$. The projected values retain their original in-plane locations, forming the base component $b_v\in\mathbb{R}^{H\times W}$.  Additional derivation details for the base projection are provided in Appendix~\ref{app:base}.

\subsection{Multiscale Residual Expansion}
\label{sec:detail}

The base component provides a compact reference to the original through-plane intensities, but a single projected value cannot preserve all $D$-dimensional intensity patterns. Simply retaining additional projections of the original through-plane vectors would still describe only the values observed at the same in-plane location across slices. Brain MRI analysis, however, often benefits from modeling local spatial context and structural boundaries across neighboring locations~\citep{Chen:2023aa}. We therefore construct the detail components from multiscale spatial descriptors before reducing across slices. To keep these components complementary to the base reference, we remove multiscale variation that is linearly associated with the sampled base scores before retaining the dominant residual variation. Below we provide the details and Fig.~\ref{fig:mvr_pipeline}(b) summarizes this construction.

\begin{figure}[tbp]
    \centering
    \includegraphics[width=\textwidth]{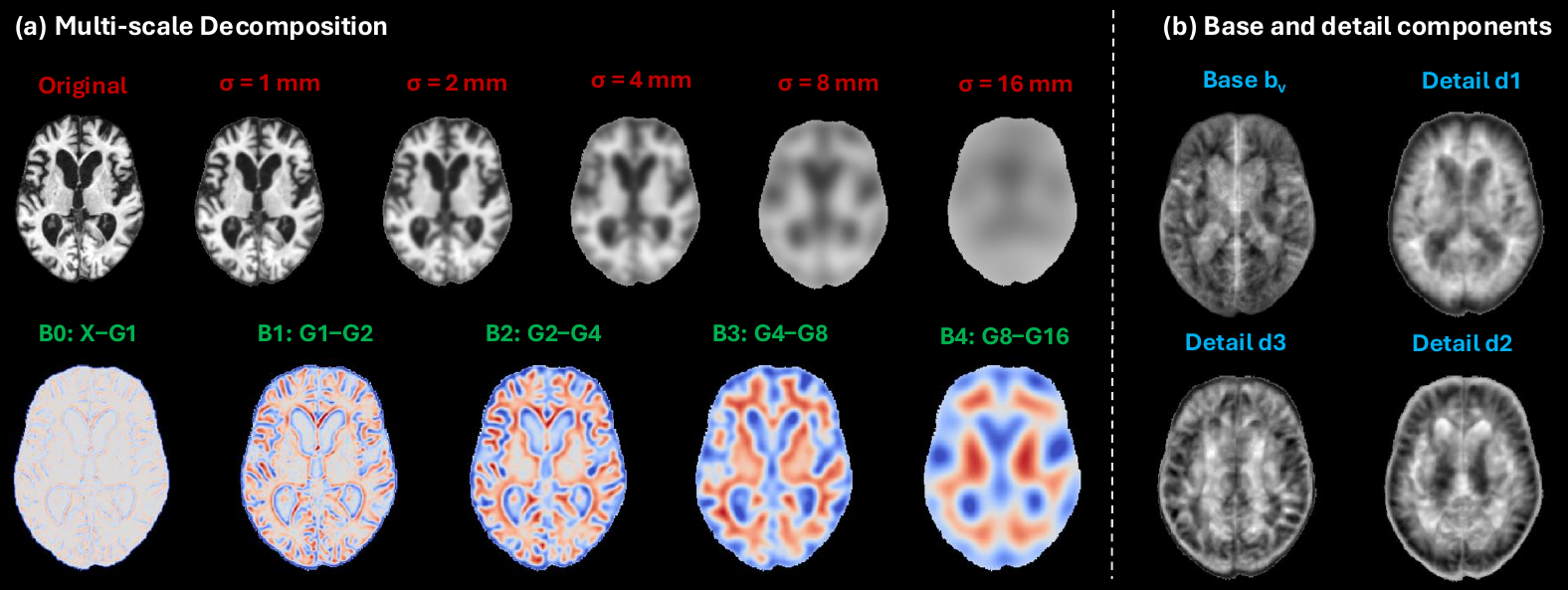}
    \caption{Visualization of the multiscale decomposition and resulting MVR components. (a) Gaussian smoothing at increasing scales produces multiscale difference components $B_0,\ldots,B_4$, capturing in-plane variation at different spatial scales. (b) The base component $b_v$ summarizes the original through-plane intensities, while the detail components capture complementary residual multiscale variation. The first four detail components are shown.}
    \label{fig:mvr_visualization}
    \vspace{-0.5em}
\end{figure}

\paragraph{Multiscale decomposition.}
To incorporate surrounding in-plane context, we construct components that describe intensity variation over neighborhoods of different sizes. Let $\mathcal{G}_{\sigma}$ denote Gaussian smoothing with standard deviation $\sigma>0$, applied independently to each 2D slice. We choose $K$ ordered smoothing scales $0<\sigma_1<\cdots<\sigma_K$. Let $L_{v,k}$ denote the smoothed volume at scale $\sigma_k$, with $L_{v,0}=I_v$, and let $B_{v,k}$ denote the difference between successive smoothing levels. These volumes retain the dimensions $H\times W\times D$ and are constructed as
\begin{equation}
\begin{aligned}
L_{v,k} &= \mathcal{G}_{\sigma_k}(I_v), && k=1,\ldots,K,\\
B_{v,k} &= L_{v,k}-L_{v,k+1}, && k=0,\ldots,K-1.
\end{aligned}
\end{equation}
These differences describe intensity variation suppressed between successive smoothing levels, providing components sensitive to different in-plane spatial scales while leaving the through-plane coordinates unchanged. We retain all $K$ difference components for detail construction. Fig.~\ref{fig:mvr_visualization}(a) visualizes the Gaussian smoothing levels and corresponding multiscale difference components for a representative slice.

To prevent difference components with larger responses from dominating the subsequent projection, we normalize each component as $\widetilde{B}_{v,k}=B_{v,k}/s_k$, where $s_k$ is its root mean square (RMS) computed over training voxels whose original intensity differs from the corresponding volume minimum. The RMS factors are estimated separately for each anatomical view and fixed thereafter.

We concatenate the normalized through-plane vectors at each location to form a multiscale descriptor $\mathbf{q}_v(h,w)\in\mathbb{R}^{KD}$:
\begin{equation}
\mathbf{q}_v(h,w)
=
\operatorname{Concat}\left(
\widetilde{B}_{v,0}(h,w,\cdot),
\ldots,
\widetilde{B}_{v,K-1}(h,w,\cdot)
\right),
\end{equation}
where each argument contains all $D$ entries in slice order. Each descriptor therefore combines in-plane context at multiple scales with the complete through-plane sequence. For each sampled through-plane vector $\mathbf{x}_i$ in $X_v$, let $\mathbf{q}_i\in\mathbb{R}^{KD}$ denote the corresponding multiscale descriptor. The sampled descriptors $\{\mathbf{q}_i\}_{i=1}^{N}$ and base scores $\{b_i\}_{i=1}^{N}$ are used together for residual fitting.

\paragraph{Residual detail extraction.}
Some multiscale variation may already be linearly associated with the sampled base scores. To remove this contribution, we first center the sampled descriptors and base scores using their training means,
$\boldsymbol{\mu}_q=\frac{1}{N}\sum_{i=1}^{N}\mathbf{q}_i\in\mathbb{R}^{KD}$
and
$\mu_b=\frac{1}{N}\sum_{i=1}^{N}b_i\in\mathbb{R}$.
We then fit a least-squares regression with coefficient vector $\boldsymbol{\beta}\in\mathbb{R}^{KD}$ to predict the centered descriptors from the centered base scores:
\begin{equation}
\begin{aligned}
\boldsymbol{\beta}
&=
\frac{
\sum_{i=1}^{N}
(b_i-\mu_b)
(\mathbf{q}_i-\boldsymbol{\mu}_q)
}{
\sum_{i=1}^{N}(b_i-\mu_b)^2
},\\
\mathbf{r}_i
&=
(\mathbf{q}_i-\boldsymbol{\mu}_q)
-
\boldsymbol{\beta}(b_i-\mu_b).
\end{aligned}
\end{equation}
The resulting residual descriptor $\mathbf{r}_i\in\mathbb{R}^{KD}$ therefore retains multiscale variation that is not linearly predicted by the base score.

We next apply PCA to the sampled residual descriptors. Stacking them as rows gives $R_v=[\mathbf{r}_1,\ldots,\mathbf{r}_N]^\top\in\mathbb{R}^{N\times KD}$, with empirical covariance matrix $C_r=R_v^\top R_v/N\in\mathbb{R}^{KD\times KD}$. We retain the mutually orthogonal unit eigenvectors $\mathbf{w}_1,\ldots,\mathbf{w}_{M-1}\in\mathbb{R}^{KD}$ corresponding to the $M-1$ largest eigenvalues of $C_r$, ordered by decreasing eigenvalue. These projection vectors capture the dominant residual multiscale variation in the sampled descriptors.

Once fitted, we construct the residual descriptor $\mathbf{r}_v(h,w)\in\mathbb{R}^{KD}$ at each in-plane location using the training means and regression coefficient. The detail-component values are then
\begin{equation}
d_{v,j}(h,w)
=
\mathbf{w}_j^\top\mathbf{r}_v(h,w),
\qquad
j=1,\ldots,M-1.
\end{equation}
The projected values retain their original in-plane locations, forming the detail components $d_{v,j}\in\mathbb{R}^{H\times W}$. Fig.~\ref{fig:mvr_visualization}(b) shows the resulting base component and the first four detail components for the same volume. The mutually orthogonal detail projection vectors therefore retain complementary directions of residual multiscale variation beyond the base reference. A detailed formulation of the residual PCA and detail projections is provided in Appendix~\ref{app:detail}.

\subsection{Encoding and classification}
\label{sec:training}

\paragraph{Foundation-model encoding.} Once fitted on the training volumes, all MVR projection vectors, regression coefficients, and associated statistics are fixed and reused for unseen volumes. The resulting base and detail components are then independently encoded by the shared frozen DINOv3 encoder. Let $\phi$ denote feature extraction using the shared frozen DINOv3 encoder. Each component is encoded independently, and the resulting features are concatenated within each view and then across the axial, coronal, and sagittal views:
\begin{equation}
\begin{aligned}
\mathbf{h}_v
&=
\operatorname{Concat}\left(
\phi(b_v),
\phi(d_{v,1}),
\ldots,
\phi(d_{v,M-1})
\right),\\
\mathbf{h}(I)
&=
\operatorname{Concat}\left(
\mathbf{h}_a,
\mathbf{h}_c,
\mathbf{h}_s
\right).
\end{aligned}
\end{equation}
The pretrained 2D foundation model weights remain fixed and the resulting $\mathbf{h}(I)$ forms the volume representation used for downstream classification.

\paragraph{Downstream classification.}
We use a linear classifier to assess the discriminative quality of $\mathbf{h}(I)$ while limiting classifier expressiveness. Each feature coordinate is standardized using its mean and standard deviation estimated from the training subjects, followed by class-balanced, L2-regularized logistic regression.

\section{Experiments}
\label{sec:experiments}

\subsection{Experimental Setup}

\paragraph{Datasets.} We evaluate MVR on two diagnostic classification tasks using three brain MRI cohorts: ADNI~\citep{ADNI} (1,027 subjects), OASIS-3~\citep{OASIS3} (550 subjects), and ABIDE~\citep{ABIDE} (1,099 subjects). ADNI and OASIS-3 are used for Alzheimer's disease (AD) versus cognitively normal (CN) classification, while ABIDE is used for autism spectrum disorder versus control classification. We further evaluate the ADNI-trained models on an external AIBL~\citep{AIBL} cohort containing 224 subjects, without AIBL-specific fitting or model selection. All methods receive the same T1-weighted volumes following N4 bias correction~\citep{N4Bias}, skull stripping, affine registration to MNI152~\citep{MNI152}, resampling to $1\,\mathrm{mm}$ isotropic resolution, and intensity normalization.

All methods use identical subject-level five-fold splits, with each fold held out for testing in turn. All data-dependent reduction and feature-standardization statistics are estimated from the corresponding training partition.

\paragraph{Metrics.} We report the area under the receiver operating characteristic curve (AUC) and balanced accuracy (BAcc). AUC measures discrimination across decision thresholds, while BAcc averages sensitivity and specificity, giving equal weight to both classes. Classification thresholds are selected using inner cross-validation on the training data and fixed before testing. Results are reported as mean $\pm$ sample standard deviation across the five held-out folds. For AIBL, we report variation across the five ADNI-fold-trained models evaluated on the same external cohort, using their ADNI-derived thresholds.

\paragraph{Comparison methods.}
We compare MVR with medically pretrained 3D foundation models, including BrainIAC~\citep{BrainIAC}, BrainMVP~\citep{BrainMVP}, and 3DINO~\citep{3DINO}. We also include MedicalNet~\citep{medicalnet} as a conventional medically pretrained 3D ResNet, together with RAPTOR~\citep{RAPTOR} and AnyMC3D~\citep{AnyMC3D}, which adapt general-purpose pretrained 2D foundation model encoders to volumetric inputs. We also evaluate several input-reduction baselines with frozen DINOv3: mean pooling, uniform slice sampling with $M=16$ fixed-interval slices per view, supervised direct projection to $M=1$ or $M=16$ components, and an adapted Eigenslices~\citep{Eigenslices} using 16 uncentered principal components per view. The supervised direct-projection baseline provides a controlled comparison in which the dimensionality and pretrained encoder are matched to MVR, but the input reduction is learned from diagnostic supervision. All pretrained encoder weights remain fixed, while method-specific trainable modules, including AnyMC3D's aggregation and the supervised input projection, are retained.

\paragraph{Implementation details.}
We use $M=16$ components per view, selected on preliminary validation and fixed for all five-fold experiments. Gaussian smoothing uses dyadic scales $\{1,2,4,8,16\}\,\mathrm{mm}$, and projection fitting uses $P=256$ sampled through-plane vectors per training subject and view. We use the frozen DINOv3 ViT-B/16 encoder~\citep{DINOv3} and concatenate the class-token representations from Transformer blocks 6 and 12. Sec.~\ref{sec:abl} and Appendix~\ref{app:extra-abl} examine sensitivity to $M$, $P$, Gaussian scales, and feature readout, while Appendix~\ref{app:implementation} describes encoder input preparation. The MVR projections and preprocessing statistics are estimated from the training subjects within each fold. For classification, we standardize features using training statistics and fit class-balanced, L2-regularized logistic regression, with regularization strength and decision threshold selected from the training data.

\begin{table}[tbp]
\centering
\caption{Classification performance under the frozen-encoder setting on ADNI, OASIS, and ABIDE. Results are mean $\pm$ sample standard deviation across five held-out folds; best and second-best means are highlighted in \textbf{bold} and \underline{underlined}. 
$^\dagger$ denotes the use of frozen DINOv3 encoder.
} 
\label{tab:main_results}

\scriptsize
\renewcommand{\arraystretch}{1.15}
\setlength{\tabcolsep}{3.8pt}

\resizebox{\linewidth}{!}{
\begin{tabular}{lcc@{\hspace{9pt}}cc@{\hspace{9pt}}cc}
\toprule
\multirow{2}{*}{\textbf{Method}}
& \multicolumn{2}{c}{\textbf{ADNI}}
& \multicolumn{2}{c}{\textbf{OASIS}}
& \multicolumn{2}{c}{\textbf{ABIDE}} \\
\cmidrule(lr){2-3}
\cmidrule(lr){4-5}
\cmidrule(lr){6-7}
& \textbf{AUC} $\uparrow$
& \textbf{BAcc} $\uparrow$
& \textbf{AUC} $\uparrow$
& \textbf{BAcc} $\uparrow$
& \textbf{AUC} $\uparrow$
& \textbf{BAcc} $\uparrow$ \\
\midrule

BrainIAC \citep{BrainIAC}
& $76.96 \pm 3.94$
& $70.34 \pm 4.66$
& $63.31 \pm 9.15$
& $58.22 \pm 6.18$
& $51.40 \pm 3.14$
& $51.48 \pm 1.35$ \\

BrainMVP \citep{BrainMVP}
& $74.38 \pm 2.70$
& $68.61 \pm 3.69$
& $67.76 \pm 5.22$
& $64.31 \pm 2.95$
& $50.41 \pm 4.11$
& $50.78 \pm 4.11$ \\

3DINO \citep{3DINO}
& $85.96 \pm 2.17$
& \underline{$78.14 \pm 2.26$}
& $67.05 \pm 2.58$
& $63.01 \pm 3.30$
& $59.30 \pm 2.43$
& $55.30 \pm 1.82$ \\

\midrule

MedicalNet \citep{medicalnet}
& \underline{$86.60 \pm 1.48$}
& {$78.10 \pm 2.00$}
& $68.35 \pm 4.99$
& $\bm{64.91 \pm 3.88}$
& $62.37 \pm 2.65$
& {$58.85 \pm 2.25$} \\

\midrule

RAPTOR \citep{RAPTOR}
& $80.26 \pm 0.94$
& $72.91 \pm 2.18$
& $64.26 \pm 5.67$
& $61.47 \pm 5.26$
& $62.95 \pm 1.79$
& $57.27 \pm 2.22$ \\

AnyMC3D$^\dagger$ \citep{AnyMC3D}
& $84.45 \pm 2.93$
& $76.90 \pm 3.81$
& $66.76 \pm 6.90$
& $62.29 \pm 5.45$
& $59.66 \pm 4.65$
& $55.16 \pm 2.05$ \\

\midrule

Eigenslices \citep{Eigenslices}
& $84.97 \pm 2.27$
& $76.89 \pm 3.43$
& \underline{$68.93 \pm 2.35$}
& $64.08 \pm 1.68$
& $61.20 \pm 3.00$
& $58.46 \pm 3.09$ \\

Mean pooling
& $81.77 \pm 3.97$
& $71.48 \pm 4.76$
& $63.48 \pm 5.60$
& $59.06 \pm 5.91$
& $55.71 \pm 3.84$
& $54.03 \pm 3.37$ \\


Uniform slice sampling ($M=16$)
& $80.87 \pm 2.14$
& $73.06 \pm 2.13$
& $62.15 \pm 2.71$
& $57.89 \pm 2.77$
& $\underline{65.99 \pm 2.88}$
& $\underline{59.56 \pm 3.34}$
\\

Direct projection ($M=1$)
& $80.08 \pm 1.15$
& $71.73 \pm 1.35$
& $61.12 \pm 4.85$
& $57.21 \pm 3.12$
& $53.99 \pm 5.09$
& $52.09 \pm 1.71$ \\

Direct projection ($M=16$)
& $82.49 \pm 1.56$
& $75.76 \pm 3.30$
& $59.25 \pm 2.95$
& $57.25 \pm 3.14$
& $59.96 \pm 1.06$
& $55.89 \pm 1.23$ \\

\midrule

\textbf{MVR (Ours)}
& $\bm{88.85 \pm 1.22}$
& $\bm{80.92 \pm 1.02}$
& $\bm{70.30 \pm 2.10}$
& \underline{$64.49 \pm 2.53$}
& $\bm{66.02 \pm 1.41}$
& $\bm{61.03 \pm 3.21}$ \\

\bottomrule
\end{tabular}
}
\end{table}

\begin{table}[ht]
\centering
\caption{ADNI-to-AIBL external evaluation without AIBL-specific fitting or model selection. Results are mean $\pm$ sample standard deviation across five ADNI-fold-trained models. Best and second-best means are highlighted in \textbf{bold} and \underline{underlined}. $^\dagger$ denotes the use of Frozen DINOv3 encoder.}
\label{tab:aibl_external}

\small
\renewcommand{\arraystretch}{1.15}
\setlength{\tabcolsep}{6pt}

\begin{tabular*}{0.7\linewidth}{@{\extracolsep{\fill}}lcc@{}}
\toprule
\textbf{Method}
& \textbf{AUC} $\uparrow$
& \textbf{BAcc} $\uparrow$ \\
\midrule

BrainIAC \citep{BrainIAC}
& $69.26 \pm 2.16$
& $60.14 \pm 1.56$ \\

BrainMVP \citep{BrainMVP}
& $78.43 \pm 0.77$
& $66.51 \pm 2.77$ \\

3DINO \citep{3DINO}
& $83.73 \pm 0.93$
& \underline{$72.19 \pm 2.71$} \\

\midrule

MedicalNet \citep{medicalnet}
& \underline{$86.21 \pm 0.25$}
& $71.45 \pm 2.63$ \\

\midrule

RAPTOR \citep{RAPTOR}
& $77.66 \pm 1.42$
& $62.95 \pm 2.60$ \\

AnyMC3D$^\dagger$ \citep{AnyMC3D}
& $80.46 \pm 0.95$
& $65.61 \pm 3.88$ \\

\midrule

Eigenslices \citep{Eigenslices}
& $85.02 \pm 0.49$
& $70.20 \pm 2.47$ \\

Mean pooling
& $78.57 \pm 0.78$
& $66.27 \pm 1.78$ \\

Uniform slice sampling ($M=16$)
& $77.95 \pm 1.76$
& $65.38 \pm 1.84$ \\

Direct projection ($M=1$)
& $74.41 \pm 4.03$
& $62.87 \pm 2.13$ \\

Direct projection ($M=16$)
& $79.25 \pm 3.23$
& $61.93 \pm 7.29$ \\

\midrule

\textbf{MVR (Ours)}
& $\bm{87.27 \pm 1.17}$
& $\bm{72.47 \pm 1.52}$ \\

\bottomrule
\end{tabular*}

\end{table}

\subsection{Results}

\paragraph{Classification performance.}
Table~\ref{tab:main_results} presents classification performance across all evaluated methods. MVR achieves the highest AUC on ADNI and OASIS at $88.85$ and $70.30$, and the highest BAcc on ADNI and ABIDE at $80.92$ and $61.03$. Compared with the medically pretrained 3D foundation models, MVR achieves higher AUC and BAcc across all three datasets. Compared with the conventional medically pretrained 3D ResNet, MedicalNet, MVR improves AUC from $86.60$ to $88.85$ on ADNI and from $68.35$ to $70.30$ on OASIS, while remaining close in OASIS BAcc ($64.49$ versus $64.91$). MVR also exceeds RAPTOR and AnyMC3D, which adapt general-purpose pretrained 2D foundation models to volumetric inputs, in both metrics across all three datasets. For the input-reduction methods, MVR achieves higher mean AUC and BAcc than Eigenslices across all three datasets. On ABIDE, uniform slice sampling achieves a mean AUC of $65.99$, close to $66.02$ for MVR, while MVR achieves higher mean BAcc ($61.03$ versus $59.56$).
At the matched $M=16$ setting, direct projection reaches AUC/BAcc of $82.49/75.76$, $59.25/57.25$, and $59.96/55.89$ on ADNI, OASIS, and ABIDE, remaining far below MVR across all six metrics. Because this baseline learns the input projection from diagnostic supervision while using the same frozen encoder and component count, the result shows that task-supervised optimization of the reduction alone does not automatically yield a stronger representation. Instead, the structured base-and-detail construction of MVR provides an effective inductive bias for presenting volumetric information to a frozen 2D encoder.


\paragraph{External generalization.}
Table~\ref{tab:aibl_external} evaluates the five ADNI-trained models on AIBL without AIBL-specific fitting or model selection. MVR achieves the highest AUC and BAcc at $87.27$ and $72.47$. Compared with the medically pretrained 3D foundation models, MVR exceeds BrainIAC, BrainMVP, and 3DINO in AUC, while its BAcc is close to 3DINO ($72.47$ versus $72.19$). Compared with the conventional medically pretrained 3D ResNet, MedicalNet, MVR improves AUC from $86.21$ to $87.27$ and BAcc from $71.45$ to $72.47$. It also exceeds RAPTOR and AnyMC3D, which adapt general-purpose pretrained 2D foundation models to volumetric inputs, in both metrics. For the input-reduction methods, Eigenslices gives the strongest comparison at $85.02/70.20$, while uniform slice sampling reaches $77.95/65.38$. Overall, the ADNI advantage of MVR is largely maintained on the external AIBL cohort.

\paragraph{Effect of encoder adaptation.} Our primary experiments intentionally keep the 2D foundation model frozen to isolate how volumetric information is presented to a pretrained 2D representation. We additionally evaluate parameter-efficient encoder adaptation using LoRA~\citep{lora} on ADNI (Appendix Table~\ref{tab:lora_results}). MVR improves slightly from 88.85/80.92 to 89.10/81.15 AUC/BAcc, whereas AnyMC3D reaches 92.93/84.56 when its encoder adaptation and slice aggregation are jointly optimized. These results indicate that the advantage of MVR should be interpreted specifically in the frozen-encoder regime rather than as a general advantage of gradient-free reduction over learned adaptation. They also suggest that learned adaptation can become advantageous when task-specific encoder optimization is permitted.
\begin{table}[tbp]
\centering
\caption{Component ablations of MVR. AUC and BAcc are reported on a $0$--$100$ scale as mean $\pm$ sample standard deviation across five held-out folds. The variant \emph{w/o base component in encoding} omits the base component from encoder inputs while retaining its use in residual construction. The variant \emph{w/o detail components} retains only the base component.}
\label{tab:component_ablation}

\scriptsize
\renewcommand{\arraystretch}{1.15}
\setlength{\tabcolsep}{3.8pt}

\resizebox{\linewidth}{!}{
\begin{tabular}{lcc@{\hspace{9pt}}cc@{\hspace{9pt}}cc}
\toprule
\multirow{2}{*}{\textbf{Variant}}
& \multicolumn{2}{c}{\textbf{ADNI}}
& \multicolumn{2}{c}{\textbf{OASIS}}
& \multicolumn{2}{c}{\textbf{ABIDE}} \\
\cmidrule(lr){2-3}
\cmidrule(lr){4-5}
\cmidrule(lr){6-7}
& \textbf{AUC} $\uparrow$
& \textbf{BAcc} $\uparrow$
& \textbf{AUC} $\uparrow$
& \textbf{BAcc} $\uparrow$
& \textbf{AUC} $\uparrow$
& \textbf{BAcc} $\uparrow$ \\
\midrule

\multicolumn{7}{l}{\textit{Multiscale construction}} \\

w/o Gaussian scale space
& $87.01 \pm 2.29$
& $78.34 \pm 3.97$
& $69.39 \pm 2.21$
& $61.11 \pm 4.61$
& $63.38 \pm 1.37$
& $59.17 \pm 3.82$ \\

\midrule
\multicolumn{7}{l}{\textit{Component composition}} \\

w/o base component in encoding
& $86.47 \pm 0.99$
& $80.69 \pm 2.35$
& $68.37 \pm 2.35$
& $62.42 \pm 2.82$
& $66.02 \pm 1.33$
& $59.21 \pm 2.40$ \\

w/o detail components
& $85.44 \pm 2.37$
& $75.68 \pm 3.18$
& $67.29 \pm 2.62$
& $60.32 \pm 3.81$
& $61.53 \pm 3.94$
& $57.94 \pm 3.18$ \\

\midrule
\multicolumn{7}{l}{\textit{Feature fusion}} \\

Mean
& $87.19 \pm 2.22$
& $80.15 \pm 3.91$
& $70.51 \pm 1.98$
& $65.71 \pm 4.46$
& $62.63 \pm 0.72$
& $58.37 \pm 1.82$ \\ 

\textbf{Concatenation (MVR)}
& $88.85 \pm 1.22$
& $80.92 \pm 1.02$
& $70.30 \pm 2.10$
& $64.49 \pm 2.53$
& $66.02 \pm 1.41$
& $61.03 \pm 3.21$ \\

\bottomrule
\end{tabular}
}
\end{table}

\begin{table}[tbp]
\centering
\caption{Effect of component count using separate encoding and concatenation across all three views. FM encodings are counted per volume. AUC and BAcc are reported on a $0$--$100$ scale as mean $\pm$ sample standard deviation across five held-out folds.}
\label{tab:component_count}

\scriptsize
\renewcommand{\arraystretch}{1.15}
\setlength{\tabcolsep}{3pt}

\resizebox{\linewidth}{!}{
\begin{tabular}{lrcc@{\hspace{7pt}}cc@{\hspace{7pt}}cc}
\toprule
\multirow{2}{*}{\textbf{Variant}}
& \multirow{2}{*}{\textbf{FM encodings}}
& \multicolumn{2}{c}{\textbf{ADNI}}
& \multicolumn{2}{c}{\textbf{OASIS}}
& \multicolumn{2}{c}{\textbf{ABIDE}} \\
\cmidrule(lr){3-4}
\cmidrule(lr){5-6}
\cmidrule(lr){7-8}
&
& \textbf{AUC} $\uparrow$
& \textbf{BAcc} $\uparrow$
& \textbf{AUC} $\uparrow$
& \textbf{BAcc} $\uparrow$
& \textbf{AUC} $\uparrow$
& \textbf{BAcc} $\uparrow$ \\
\midrule

$M=3$
& $9$
& $86.98 \pm 2.18$
& $77.76 \pm 1.66$
& $70.72 \pm 2.46$
& $65.22 \pm 2.67$
& $62.57 \pm 2.36$
& $59.18 \pm 3.68$ \\

$M=5$
& $15$
& $87.67 \pm 1.96$
& $80.64 \pm 0.44$
& $70.65 \pm 2.92$
& $63.16 \pm 1.79$
& $64.84 \pm 1.90$
& $57.66 \pm 3.49$ \\

$M=8$
& $24$
& $87.89 \pm 1.27$
& $80.19 \pm 1.53$
& $70.53 \pm 2.50$
& $63.51 \pm 5.32$
& $64.80 \pm 0.86$
& $59.33 \pm 2.22$ \\

$M=16$ (default)
& $48$
& $88.85 \pm 1.22$
& $80.92 \pm 1.02$
& $70.30 \pm 2.10$
& $64.49 \pm 2.53$
& $66.02 \pm 1.41$
& $61.03 \pm 3.21$ \\

$M=32$
& $96$
& $88.28 \pm 1.37$
& $80.01 \pm 0.99$
& $70.41 \pm 2.02$
& $63.56 \pm 2.88$
& $65.41 \pm 1.51$
& $58.04 \pm 3.56$ \\

\bottomrule
\end{tabular}
}
\end{table}

\subsection{Ablation Study}
\label{sec:abl}

Table~\ref{tab:component_ablation} examines the contributions of multiscale construction, component composition, and feature fusion. Below we provide detailed analysis.
 
\paragraph{Multiscale construction.}
Removing the Gaussian scale space while retaining the remaining MVR construction reduces AUC by $1.84$ points on ADNI, $0.91$ on OASIS, and $2.64$ on ABIDE. The corresponding BAcc drops are $2.58$, $3.38$, and $1.86$ points. The consistent degradation across all three datasets supports incorporating multiscale in-plane variation before through-plane reduction.

\paragraph{Component composition.}
Retaining only the base component reduces AUC by $3.41$ points on ADNI, $3.01$ on OASIS, and $4.49$ on ABIDE, while BAcc decreases by $5.24$, $4.17$, and $3.09$ points. This shows that the base component alone is insufficient to form the complete representation. Removing the base component from the encoder inputs while retaining it for residual construction also lowers both metrics on ADNI and OASIS. On ABIDE, mean AUC remains unchanged at the reported precision, while BAcc decreases from $61.03$ to $59.21$. The encoded base component therefore provides a complementary overall contribution, while the detail components deliver larger and more consistent gains across all three datasets.

\paragraph{Feature fusion.}
Full concatenation performs best on ADNI and ABIDE, improving AUC over mean fusion by $1.66$ and $3.39$ points and BAcc by $0.77$ and $2.66$ points. On OASIS, mean fusion performs slightly better, reaching $70.51$ AUC and $65.71$ BAcc compared with $70.30$ and $64.49$ for concatenation. These results show that preserving separate component features is beneficial on ADNI and ABIDE, while averaging remains competitive on OASIS.

\paragraph{Component count.}
Table~\ref{tab:component_count} examines the number of MVR components retained per anatomical view. On ADNI, performance generally improves with $M$ and peaks at $M=16$, reaching $88.85$ AUC and $80.92$ BAcc. ABIDE shows a similar pattern, with $M=16$ achieving the highest AUC and BAcc of $66.02$ and $61.03$. OASIS is less sensitive to component count, with AUC remaining between $70.30$ and $70.72$ across all tested values and the highest AUC/BAcc obtained at $M=3$. Increasing $M$ from 16 to 32 provides no consistent improvement across datasets and metrics.

\section{Conclusion}
We introduced Multiscale Volumetric Reduction (MVR), a data-adaptive framework that departs from the conventional encode-then-integrate paradigm by reducing volumetric information before foundation-model encoding. MVR constructs complementary base and multiscale detail components that preserve through-plane structure while incorporating spatial context, enabling a frozen 2D foundation model to represent a 3D brain MRI volume using a compact set of dense 2D inputs. Across multiple brain-MRI cohorts and diagnostic tasks, MVR achieves strong and consistent performance and transfers well to an external cohort. These results support reduction before encoding as a viable alternative to post-encoding volumetric integration for adapting frozen 2D foundation models to brain MRI.

\bibliography{iclr2027_conference}
\bibliographystyle{iclr2027_conference}

\appendix
\section{Appendix}

\subsection{Detailed Formulation of the MVR Projections}

We provide the detailed formulation of the base and detail projections used in MVR, including their PCA objectives and eigendecomposition solutions.

\subsubsection{Base Projection}
\label{app:base}

Let $X_v=[\mathbf{x}_1,\ldots,\mathbf{x}_N]^\top\in\mathbb{R}^{N\times D}$ contain the sampled through-plane vectors for anatomical view $v$, following the notation in the main paper Sec.\ref{sec:base}. We want to find a unit projection vector $\mathbf{u}\in\mathbb{R}^{D}$ that maximizes the average squared projection of these vectors:
\begin{equation}
\mathbf{u}_b
=
\arg\max_{\|\mathbf{u}\|_2=1}
\frac{1}{N}\sum_{i=1}^{N}
\left(\mathbf{x}_i^\top\mathbf{u}\right)^2.
\label{eq:app_base_objective}
\end{equation}
The objective can be rewritten as
\begin{equation}
\frac{1}{N}\sum_{i=1}^{N}
\left(\mathbf{x}_i^\top\mathbf{u}\right)^2
=
\mathbf{u}^\top
\left(\frac{1}{N}X_v^\top X_v\right)
\mathbf{u}
=
\mathbf{u}^\top M_b\mathbf{u},
\end{equation}
where
\begin{equation}
M_b=\frac{1}{N}X_v^\top X_v
\end{equation}
is the uncentered second-moment matrix.

Since $M_b$ is symmetric and positive semidefinite, we compute its eigendecomposition
\begin{equation}
M_b
=
U_b\Lambda_b U_b^\top,
\end{equation}
where the eigenvalues in $\Lambda_b$ are ordered from largest to smallest. The objective in Eq.~\ref{eq:app_base_objective} is maximized by the unit eigenvector corresponding to the largest eigenvalue. We therefore select this eigenvector as the base projection vector $\mathbf{u}_b$.

The use of uncentered PCA is motivated by the role of the base component as an original-intensity reference. Let $\boldsymbol{\mu}_x=\frac{1}{N}\sum_{i=1}^{N}\mathbf{x}_i$ denote the mean sampled through-plane vector and let
\begin{equation}
C_x
=
\frac{1}{N}\sum_{i=1}^{N}
(\mathbf{x}_i-\boldsymbol{\mu}_x)
(\mathbf{x}_i-\boldsymbol{\mu}_x)^\top
\end{equation}
denote the centered covariance matrix. Expanding the covariance gives
\begin{equation}
C_x
=
M_b-\boldsymbol{\mu}_x\boldsymbol{\mu}_x^\top,
\end{equation}
or equivalently
\begin{equation}
M_b
=
C_x+\boldsymbol{\mu}_x\boldsymbol{\mu}_x^\top.
\label{eq:app_second_moment_decomposition}
\end{equation}
Thus, centered PCA removes the average through-plane intensity structure before estimating its principal directions, whereas the uncentered second-moment matrix retains this structure. This is consistent with using the base component as an original-intensity reference.

Applying the fitted projection vector to the sampled through-plane vectors gives the sampled base scores
\begin{equation}
\mathbf{b}
=
X_v\mathbf{u}_b
=
[b_1,\ldots,b_N]^\top,
\qquad
b_i=\mathbf{x}_i^\top\mathbf{u}_b.
\end{equation}
The same projection is applied at each in-plane location of a volume to construct the base component:
\begin{equation}
b_v(h,w)
=
\mathbf{u}_b^\top\mathbf{x}_v(h,w).
\end{equation}

\subsubsection{Detail Projection}
\label{app:detail}

Following the residualization defined in the main paper Sec. \ref{sec:detail}, let $\mathbf{r}_i\in\mathbb{R}^{KD}$ denote the residual descriptor associated with the $i$-th sampled multiscale descriptor. Stacking the residual descriptors as rows gives $R_v=[\mathbf{r}_1,\ldots,\mathbf{r}_N]^\top\in\mathbb{R}^{N\times KD}$, with covariance matrix
\begin{equation}
C_r=\frac{1}{N}R_v^\top R_v.
\end{equation}

The detail projections retain the directions with the largest residual variance. For the first detail projection, we solve
\begin{equation}
\mathbf{w}_1
=
\arg\max_{\|\mathbf{w}\|_2=1}
\mathbf{w}^\top C_r\mathbf{w}.
\end{equation}
Subsequent projection vectors are obtained under orthogonality constraints:
\begin{equation}
\mathbf{w}_j
=
\arg\max_{\|\mathbf{w}\|_2=1}
\mathbf{w}^\top C_r\mathbf{w},
\qquad
\mathbf{w}^\top\mathbf{w}_{\ell}=0
\quad
\text{for } \ell<j.
\end{equation}

Since $C_r$ is symmetric and positive semidefinite, we compute its eigendecomposition
\begin{equation}
C_r=U_r\Lambda_rU_r^\top,
\end{equation}
where the eigenvalues in $\Lambda_r$ are ordered from largest to smallest. The $M-1$ detail projection vectors are therefore the unit eigenvectors corresponding to the $M-1$ largest eigenvalues:
\begin{equation}
[\mathbf{w}_1,\ldots,\mathbf{w}_{M-1}]
=
U_r[:,1{:}M-1].
\end{equation}
The selected projection vectors capture the dominant residual multiscale variation and are mutually orthogonal in the residual PCA space. Applying $\mathbf{w}_j$ to the residual descriptor $\mathbf{r}_v(h,w)$ gives the corresponding detail-component value, as defined in the main text.

\subsection{Ablation}
\label{app:extra-abl}
\begin{table}[ht]
\centering
\caption{Effect of the number of sampled through-plane vectors per training subject on ADNI. AUC and BAcc are reported on a $0$--$100$ scale as mean $\pm$ sample standard deviation across five held-out folds. Best means are in \textbf{bold} and second-best means are \underline{underlined}.}
\label{tab:sampling_ablation}

\scriptsize
\renewcommand{\arraystretch}{1.15}
\setlength{\tabcolsep}{3pt}

\begin{tabular}{ccc}
\toprule
\textbf{$P$}
& \textbf{AUC} $\uparrow$
& \textbf{BAcc} $\uparrow$ \\
\midrule

$64$
& $88.33 \pm 1.63$
& $80.84 \pm 1.51$ \\

$128$
& $88.46 \pm 1.45$
& $79.99 \pm 1.51$ \\

$256$
& $\bm{88.85 \pm 1.22}$
& \underline{$80.92 \pm 1.02$} \\

$512$
& \underline{$88.80 \pm 1.37$}
& $\bm{81.68 \pm 0.87}$ \\
\midrule

All
& $88.62 \pm 1.46$
& $80.59 \pm 1.52$ \\

\bottomrule
\end{tabular}
\end{table}

\paragraph{Sampling sensitivity.} Table~\ref{tab:sampling_ablation} examines the number of through-plane vectors sampled per training subject for fitting the MVR projections. Performance is relatively stable across the tested values, with AUC ranging from $88.33$ to $88.85$ and BAcc from $79.99$ to $81.68$. The default setting $P=256$ achieves the highest AUC ($88.85$), while $P=512$ gives the highest BAcc ($81.68$). Using all available vectors does not improve over the sampled settings, indicating that a moderate number of sampled vectors is sufficient for estimating the shared projections.

\begin{table}[ht]
\centering
\caption{Effect of anatomical-view composition on classification performance. AUC and BAcc are reported on a $0$--$100$ scale. Best results are highlighted in bold and second-best results are underlined.}
\label{tab:view_ablation}

\scriptsize
\renewcommand{\arraystretch}{1.15}
\setlength{\tabcolsep}{3.8pt}

\begin{tabular}{lcc@{\hspace{9pt}}cc@{\hspace{9pt}}cc}
\toprule
\multirow{2}{*}{\textbf{Retained views}}
& \multicolumn{2}{c}{\textbf{ADNI}}
& \multicolumn{2}{c}{\textbf{OASIS}}
& \multicolumn{2}{c}{\textbf{ABIDE}} \\
\cmidrule(lr){2-3}
\cmidrule(lr){4-5}
\cmidrule(lr){6-7}
& \textbf{AUC} $\uparrow$
& \textbf{BAcc} $\uparrow$
& \textbf{AUC} $\uparrow$
& \textbf{BAcc} $\uparrow$
& \textbf{AUC} $\uparrow$
& \textbf{BAcc} $\uparrow$ \\
\midrule

Axial
& $86.80$
& $79.22$
& $68.80$
& $62.79$
& {$65.86$}
& {$60.77$} \\

Coronal
& $85.36$
& $77.52$
& $69.94$
& $64.32$
& $61.76$
& $58.58$ \\

Sagittal
& $85.77$
& $76.60$
& $67.88$
& $62.43$
& $62.71$
& $58.52$ \\
\midrule

Axial + coronal
& $87.90$
& $80.05$
& $69.77$
& $62.17$
& $65.25$
& $\bm{61.47}$ \\

Axial + sagittal
& \underline{$88.48$}
& \underline{$80.80$}
& $69.47$
& $63.15$
& $\underline{65.95}$
& $59.92$ \\

Coronal + sagittal
& $87.20$
& $79.89$
& \underline{$70.16$}
& \underline{$64.34$}
& $63.27$
& $59.71$ \\
\midrule

All three
& $\bm{88.85}$
& $\bm{80.92}$
& $\bm{70.30}$
& $\bm{64.49}$
& $\bm{66.02}$
& $\underline{61.03}$ \\

\bottomrule
\end{tabular}
\end{table}

\paragraph{Anatomical-view composition.} Table~\ref{tab:view_ablation} evaluates the contribution of the axial, coronal, and sagittal views individually and in combination. Using all three views gives the highest AUC and BAcc on ADNI ($88.85/80.92$) and OASIS ($70.30/64.49$), supporting the use of complementary information across anatomical orientations. On ABIDE, using all three views gives the highest AUC ($66.02$), while axial+coronal gives the highest BAcc ($61.47$). Nevertheless, the three-view representation remains competitive on both metrics, suggesting that the relative contribution of each anatomical view is dataset dependent.

\begin{table}[!ht]
\centering
\caption{Classification performance with LoRA adaptation on ADNI. AUC and BAcc are reported on a $0$--$100$ scale as mean $\pm$ sample standard deviation across five held-out folds. Best means are in \textbf{bold} and second-best means are \underline{underlined}.}
\label{tab:lora_results}

\scriptsize
\renewcommand{\arraystretch}{1.15}
\setlength{\tabcolsep}{3pt}

\begin{tabular}{lcc}
\toprule
\textbf{Method}
& \textbf{AUC} $\uparrow$
& \textbf{BAcc} $\uparrow$ \\
\midrule

BrainIAC
& $79.65 \pm 2.29$
& $72.01 \pm 1.20$ \\

BrainMVP 
& $85.24 \pm 3.38$
& $75.25 \pm 4.80$ \\

3DINO 
& $87.01 \pm 1.60$
& $79.10 \pm 0.94$ \\

AnyMC3D
& $\bm{92.93 \pm 3.14}$
& $\bm{84.56 \pm 4.28}$ \\
\midrule

MVR (Ours)
& \underline{$89.10 \pm 1.16$}
& \underline{$81.15 \pm 2.59$} \\

\bottomrule
\end{tabular}
\end{table}

\begin{table}[!ht]
\centering
\caption{Sensitivity to the Gaussian scale configuration on ADNI. AUC and BAcc are reported on a $0$--$100$ scale as mean $\pm$ sample standard deviation across five held-out folds. Best means are in \textbf{bold} and second-best means are \underline{underlined}.}
\label{tab:scale_ablation}

\scriptsize
\renewcommand{\arraystretch}{1.15}
\setlength{\tabcolsep}{5pt}

\begin{tabular}{lcc}
\toprule
\textbf{Gaussian scales (mm)}
& \textbf{AUC} $\uparrow$
& \textbf{BAcc} $\uparrow$ \\
\midrule

$\{1,2,4,8,16\}$
& \underline{$88.85 \pm 1.22$}
& $\bm{80.92 \pm 1.02}$ \\

$\{1,\sqrt{2},2,2\sqrt{2},4,4\sqrt{2},8,8\sqrt{2},16\}$
& $\bm{88.93 \pm 1.58}$
& $\bm{80.92 \pm 2.34}$ \\

\bottomrule
\end{tabular}
\end{table}

\paragraph{Gaussian-scale sensitivity.} Table~\ref{tab:scale_ablation} compares the default five-scale configuration with a denser nine-scale configuration spanning the same $1$--$16\,\mathrm{mm}$ range. Increasing the sampling density of the Gaussian scale space changes the mean AUC only marginally, from $88.85$ to $88.93$, while the mean BAcc remains unchanged at $80.92$. This suggests that MVR is not sensitive to the precise density of Gaussian scales within the evaluated range.

\paragraph{LoRA adaptation.} Table~\ref{tab:lora_results} evaluates parameter-efficient encoder adaptation using LoRA with a common rank of $r=8$ across all methods. MVR reaches $89.10$ AUC and $81.15$ BAcc, outperforming the medically pretrained 3D encoders. AnyMC3D achieves the highest performance, with $92.93$ AUC and $84.56$ BAcc. This setting is outside the primary design of MVR, which constructs the volumetric representation for a frozen DINOv3 encoder without gradient-based encoder adaptation. By contrast, AnyMC3D is designed to jointly optimize trainable encoder adaptation and slice aggregation for volumetric classification. The LoRA results therefore show that MVR remains competitive when encoder adaptation is introduced, while its primary setting is the use of a frozen pretrained 2D foundation model.

\begin{table}[t]
\centering
\caption{Feature-layer ablation on ADNI. AUC and BAcc are reported on a $0$--$100$ scale as mean $\pm$ sample standard deviation across five held-out folds. Best results are highlighted in \textbf{bold}.}
\label{tab:feature_layer_ablation}
\small
\begin{tabular}{lcc}
\toprule
Representation & AUC $\uparrow$ & BAcc $\uparrow$ \\
\midrule
Block 6 only
& $86.83 \pm 1.61$
& $78.61 \pm 0.62$ \\

Block 12 only
& $88.31 \pm 1.32$
& $80.62 \pm 1.99$ \\

Block 6 + Block 12
& $\mathbf{88.85 \pm 1.22}$
& $\mathbf{80.92 \pm 1.02}$ \\
\bottomrule
\end{tabular}
\end{table}

\paragraph{Feature-layer ablation.} Table~\ref{tab:feature_layer_ablation} compares representations extracted from intermediate and final DINOv3 blocks on ADNI. Block 12 provides a stronger single-layer representation than Block 6, improving AUC by 1.48 points and BAcc by 2.01 points. Concatenating features from Blocks 6 and 12 further improves performance to 88.85 AUC and 80.92 BAcc, corresponding to gains of 0.54 AUC points and 0.30 BAcc points over Block 12 alone. These results suggest that the final block contains the strongest task-relevant representation, while the intermediate block contributes complementary information. The additional gain is modest relative to the doubled classifier input dimension, although concatenating the two blocks requires no additional DINOv3 forward passes.

\subsection{Implementation}
\label{app:implementation}

\paragraph{Component scaling and encoder input preparation.}
For each anatomical view $v$ and component $m$, the projected values are reshaped to form a 2D component $Z_{v,m}\in\mathbb{R}^{H_v\times W_v}$. Foreground locations are identified from the constant minimum-valued background of each normalized volume, without requiring an external brain mask or segmentation. All scaling statistics are estimated from the corresponding training fold. Let $q^{1}_{v,m}$ and $q^{99}_{v,m}$ denote the 1st and 99th percentiles of the training values for component $m$. We define
\begin{equation}
c_{v,m}=\frac{q^{1}_{v,m}+q^{99}_{v,m}}{2},
\qquad
s_{v,m}=\max\left(
\frac{q^{99}_{v,m}-q^{1}_{v,m}}{2},
10^{-8}
\right).
\end{equation}
Each foreground value is then mapped to $[0,1]$ as
\begin{equation}
\widetilde Z_{v,m}(p)
=
\frac{1}{2}
\left[
\operatorname{clip}
\left(
\frac{Z_{v,m}(p)-c_{v,m}}
{s_{v,m}L_v},
-1,1
\right)
+1
\right],
\end{equation}
where $L_v$ is the 99th percentile of the absolute standardized responses estimated from the training fold for view $v$. Background locations are set to zero. Each resulting component is replicated across three channels, resized to $256\times256$ using antialiased bilinear interpolation, and normalized using the ImageNet channel statistics used by DINOv3. All scaling and clipping parameters are fixed before application to validation and test subjects.

\paragraph{Uniform Slice Sampling.} 
For the uniform-sampling comparison, we select $M=16$ fixed slice positions independently for each anatomical view and outer fold. For each training subject $s$, foreground is identified directly from the constant minimum-valued background of the normalized volume. A slice at depth $d$ is considered occupied if it contains at least one foreground voxel. Let $a_{s,v}$ and $b_{s,v}$ denote the first and one-past-last occupied slice for subject $s$ in view $v$. We obtain a fold-level foreground interval from the median boundaries of the training subjects,
\begin{equation}
a_v=\operatorname{round}\!\left(\operatorname{median}_{s\in\mathcal{T}}a_{s,v}\right),
\qquad
b_v=\operatorname{round}\!\left(\operatorname{median}_{s\in\mathcal{T}}b_{s,v}\right).
\end{equation}
The interval $[a_v,b_v)$ is divided into $M$ equal-width bins, and the slice nearest the center of each bin is selected:
\begin{equation}
i_{v,m}
=
\left\lfloor
a_v+
\left(m+\frac{1}{2}\right)
\frac{b_v-a_v}{M}
\right\rfloor,
\qquad
m=0,\ldots,M-1.
\end{equation}
The resulting indices are fixed and applied to every training and held-out subject in the corresponding fold, so slice position $m$ represents approximately the same anatomical location across subjects. Each selected slice is prepared using the same training-fold intensity scaling and DINOv3 input preprocessing as the MVR components, encoded independently by the frozen encoder, and concatenated across slices and views.

\end{document}